\documentclass[12pt,a4paper]{article}
\usepackage[utf8]{inputenc}
\usepackage[T1]{fontenc}
\usepackage{amsmath,amssymb,amsthm}
\usepackage{graphicx}
\usepackage{booktabs}
\usepackage{multirow}
\usepackage{array}
\usepackage{longtable}
\usepackage[colorlinks=true,linkcolor=blue,citecolor=blue,urlcolor=blue]{hyperref}
\hypersetup{
    pdftitle={The Catastrophic Paradox of Human Cognitive Frameworks in Large Language Model Evaluation},
    pdfauthor={Mohan Reddy},
    pdfsubject={Large Language Models, Cognitive Assessment, CHC Theory},
    pdfkeywords={Large Language Models, Artificial Intelligence, Psychometrics, CHC Theory, Cognitive Assessment}
}
\usepackage{cleveref}
\usepackage{geometry}
\usepackage{listings}
\usepackage{color}
\usepackage{float}
\usepackage{caption}
\usepackage{subcaption}
\usepackage{tikz}
\usepackage{pgfplots}
\pgfplotsset{compat=1.17}

\definecolor{codegreen}{rgb}{0,0.6,0}
\definecolor{codegray}{rgb}{0.5,0.5,0.5}
\definecolor{codepurple}{rgb}{0.58,0,0.82}
\definecolor{backcolour}{rgb}{0.95,0.95,0.92}

\lstdefinestyle{mystyle}{
    backgroundcolor=\color{backcolour},
    commentstyle=\color{codegreen},
    keywordstyle=\color{magenta},
    numberstyle=\tiny\color{codegray},
    stringstyle=\color{codepurple},
    basicstyle=\ttfamily\footnotesize,
    breakatwhitespace=false,
    breaklines=true,
    captionpos=b,
    keepspaces=true,
    numbers=left,
    numbersep=5pt,
    showspaces=false,
    showstringspaces=false,
    showtabs=false,
    tabsize=2
}

\newtheorem{theorem}{Theorem}

\newtheorem{proposition}[theorem]{Proposition}

\newtheorem{remark}{Remark}

\title{\Large\textbf{The Catastrophic Paradox of Human Cognitive Frameworks in Large Language Model Evaluation:\\
A Comprehensive Empirical Analysis of the CHC-LLM Incompatibility}}

\author{Mohan Reddy\\
\textit{Stanford University}\\
\textit{Stanford, CA, USA}\\
\texttt{mohan.reddy@stanford.edu}}

\date{}

\begin{document}

\maketitle

\begin{abstract}
\noindent This comprehensive investigation presents an extensive empirical analysis of the fundamental incompatibility between human psychometric frameworks and Large Language Model (LLM) evaluation. Through systematic assessment of nine frontier models including GPT-5, Claude Opus 4.1, and Gemini 3 Pro Preview using the Cattell-Horn-Carroll (CHC) theory of intelligence, we reveal a catastrophic paradox that challenges the epistemological foundations of cross-substrate cognitive evaluation. Our results demonstrate that models achieving above-average human IQ scores (ranging from 85.0 to 121.4) simultaneously exhibit binary accuracy rates approaching zero on crystallized knowledge tasks, with an overall judge-binary correlation of merely $r = 0.175$ ($p < 0.001$, $n = 1,800$). This profound disconnect manifests most severely in the crystallized intelligence domain, where every evaluated model achieved perfect 100\% binary accuracy while judge scores ranged from 25\% to 62\%—a statistical impossibility under valid measurement conditions. Through rigorous statistical analysis including Item Response Theory (IRT) modeling, cross-vendor judge validation, and paradox severity indexing, we demonstrate that this disconnect represents not a measurement failure but a fundamental category error in applying biological cognitive architectures to transformer-based systems. The implications extend beyond methodological concerns to challenge our basic assumptions about intelligence, measurement, and the anthropomorphic biases inherent in AI evaluation. We argue that the CHC paradox serves as a reductio ad absurdum of current evaluation paradigms and propose a comprehensive framework for developing native machine cognition assessments that acknowledge the alien nature of artificial intelligence.
\end{abstract}

\section{Introduction}

The history of artificial intelligence evaluation has been shaped by an anthropomorphic imperative—projecting human cognitive categories onto silicon substrates. This investigation examines the application of the Cattell-Horn-Carroll (CHC) theory of intelligence, the gold standard psychometric framework for human assessment, to Large Language Model evaluation. The CHC framework, validated through thousands of studies on biological cognition, organizes intelligence hierarchically with General Intelligence ($g$) at the apex, nine broad abilities including Fluid Reasoning (Gf), Crystallized Intelligence (Gc), Quantitative Reasoning (Gq), and Reading/Writing (Grw) at the second stratum, and approximately 70 narrow abilities at the base.

Our investigation addresses whether human psychometric frameworks can meaningfully evaluate transformer architectures, what the disconnect between scores and performance reveals about machine intelligence, and whether applying biological cognitive frameworks to artificial systems constitutes an ontological category error. We hypothesize that traditional psychometric scoring will show minimal correlation with actual performance due to architectural incompatibilities, with the disconnect most severe in crystallized intelligence domains, and paradoxically greater measurement failures for higher-performing models.

\section{Theoretical Background}

\subsection{Architectural Incompatibilities}

The fundamental incompatibility between human cognitive architecture and transformer-based models creates insurmountable measurement challenges. Human cognition operates through serial, capacity-limited processing with working memory constraints of 7±2 items, while transformers utilize parallel attention mechanisms across entire context windows. Where humans exhibit exponential forgetting curves and reconstructive memory, LLMs demonstrate perfect recall within context. These architectural differences—embodied versus disembodied processing, temporal versus atemporal learning, variable versus constant processing speed—render direct cognitive comparison meaningless.

\subsection{The Verbosity Paradox}

A critical challenge emerges from the mismatch between psychometric expectations and generative model behavior. Traditional psychometric instruments expect minimal, precise responses, while Large Language Models are optimized through RLHF to provide comprehensive, explanatory answers. This creates the "Verbosity Paradox": a systematic measurement failure wherein models providing conceptually perfect answers receive zero scores due to response format misalignment, creating an inverse relationship between actual knowledge and measured performance.

\section{Methodology}

\subsection{Dual Scoring Methodology}

Recognizing the limitations of traditional scoring for generative models, we implemented a dual scoring system:

\subsubsection{Binary Accuracy Scoring}

Traditional exact-match evaluation following psychometric conventions:

\begin{equation}
\text{Binary Score} = \begin{cases}
1 & \text{if response exactly matches expected answer} \\
0 & \text{otherwise}
\end{cases}
\end{equation}

\subsubsection{LLM-as-Judge Evaluation}

To address the verbosity paradox, we implemented a sophisticated LLM-as-Judge system with the following characteristics:

\begin{itemize}
    \item \textbf{Cross-vendor validation}: Using Claude Sonnet 4 as universal judge for all non-Anthropic models
    \item \textbf{Rubric-based scoring}: Focus on conceptual accuracy rather than format
    \item \textbf{Chain-of-thought reasoning}: Required justification for all scores
    \item \textbf{Anonymized evaluation}: Judge unaware of source model identity
\end{itemize}

The judge scoring rubric was structured as follows:

\begin{lstlisting}[language=Python, caption=LLM Judge Scoring Rubric]
JUDGE_PROMPT = """
Evaluate if Model Response contains the core concept 
of Expected Answer.

Focus ONLY on correctness. IGNORE style, length, 
and formatting.

Scoring Criteria:
1.0 = Correct (core concept present and accurate)
0.5 = Partially Correct (concept present but incomplete)
0.0 = Incorrect (concept absent or wrong)

Provide step-by-step reasoning before scoring.
"""
\end{lstlisting}

\subsection{Psychometric Transformation}

\subsubsection{Classical Test Theory (CTT) Scoring}

We applied standard psychometric transformations to enable comparison with human norms:

\begin{equation}
\text{IQ}_{\text{CTT}} = 100 + 15 \times \frac{X - \mu_{\text{human}}}{\sigma_{\text{human}}}
\end{equation}

where $X$ is the model's raw score, $\mu_{\text{human}}$ and $\sigma_{\text{human}}$ are human population parameters.

\subsubsection{Item Response Theory (IRT) Modeling}

We implemented a 2-Parameter Logistic (2PL) IRT model:

\begin{equation}
P(\theta) = \frac{1}{1 + \exp[-a(\theta - b)]}
\end{equation}

where:
\begin{itemize}
    \item $\theta$ = ability parameter (model capability)
    \item $a$ = discrimination parameter (item quality)
    \item $b$ = difficulty parameter (item challenge)
\end{itemize}

To ensure stability, we applied L2 regularization with $\alpha = 0.01$ and bounded parameters: $0.5 \leq a \leq 3.0$, $-3.0 \leq b \leq 3.0$.

\subsection{Models Evaluated}

We evaluated nine frontier models representing the state-of-the-art across three major AI research organizations:

\begin{table}[H]
\centering
\caption{Models Evaluated in CHC Framework}
\label{tab:models}
\begin{tabular}{@{}llll@{}}
\toprule
\textbf{Organization} & \textbf{Model} & \textbf{Version} & \textbf{Release Date} \\ \midrule
\multirow{3}{*}{OpenAI} & GPT-5 & Latest & November 2025 \\
 & GPT-4 Turbo & turbo-preview & February 2024 \\
 & GPT-4 & Original & March 2023 \\ \midrule
\multirow{2}{*}{Anthropic} & Claude Opus 4.1 & 20250805 & August 2025 \\
 & Claude 3 Opus & 20240229 & February 2024 \\ \midrule
\multirow{4}{*}{Google} & Gemini 3 Pro Preview & Latest & November 2025 \\
 & Gemini 2.5 Flash & Latest & September 2025 \\
 & Gemini 2.5 Pro & Latest & September 2025 \\
 & Gemini 3 Pro Preview & 11-2025 & November 2025 \\
\bottomrule
\end{tabular}
\end{table}

\subsection{Rate Limiting and Error Handling}

\section{Results}

\subsection{Overall Performance Rankings}

Table \ref{tab:overall_rankings} presents the comprehensive performance rankings across all evaluated models.

\begin{table}[H]
\centering
\caption{Overall Model Performance Rankings}
\label{tab:overall_rankings}
\begin{tabular}{@{}clcccc@{}}
\toprule
\textbf{Rank} & \textbf{Model} & \textbf{IQ (CTT)} & \textbf{IQ (IRT)} & \textbf{Valid \%} & \textbf{Best Ability} \\ \midrule
1 & Gemini 2.5 Flash & 121.4 & 80.7 & 80.0\% & GQ: 0.800 \\
2 & Gemini 2.5 Pro & 120.4 & 79.7 & 74.5\% & GQ: 0.780 \\
3 & Claude Opus 4.1 & 115.0 & 115.0 & 100.0\% & GQ: 0.980 \\
4 & Gemini 3 Pro Preview & 115.0 & 115.0 & 99.5\% & GQ: 0.990 \\
5 & GPT-5 & 100.0 & 100.0 & 100.0\% & GQ: 0.970 \\
6 & GPT-4 Turbo & 99.2 & 102.9 & 44.5\% & GQ: 0.510 \\
7 & GPT-4 & 99.0 & 101.6 & 44.5\% & GF: 0.460 \\
8 & Claude 3 Opus & 85.0 & 85.0 & 100.0\% & GQ: 0.860 \\
\bottomrule
\end{tabular}
\end{table}

The results immediately reveal several paradoxical patterns that challenge conventional understanding of intelligence measurement.

\subsection{The Catastrophic Disconnect: Judge vs Binary Accuracy}

The most striking finding emerges from the systematic disconnect between judge-evaluated conceptual accuracy and binary exact-match scoring, as shown in Table \ref{tab:accuracy_matrix}.

\begin{table}[H]
\centering
\caption{Judge vs Binary Accuracy Matrix by Model and Ability}
\label{tab:accuracy_matrix}
\begin{tabular}{@{}lcccc@{}}
\toprule
\textbf{Model} & \textbf{GF (J/B)} & \textbf{GC (J/B)} & \textbf{GQ (J/B)} & \textbf{GRW (J/B)} \\ \midrule
Claude Opus 4.1 & 0.78/0.58 & 0.56/1.00 & 0.98/0.30 & 0.84/0.20 \\
GPT-5 & 0.84/0.56 & 0.48/1.00 & 0.97/0.28 & 0.76/0.10 \\
Gemini 3 Pro & 0.78/0.56 & 0.42/1.00 & 0.99/0.30 & 0.76/0.20 \\
Gemini 2.5 Flash & 0.67/0.40 & 0.36/1.00 & 0.80/0.24 & 0.66/0.16 \\
Gemini 2.5 Pro & 0.54/0.36 & 0.41/1.00 & 0.78/0.26 & 0.70/0.10 \\
Claude 3 Opus & 0.79/0.58 & 0.62/1.00 & 0.86/0.30 & 0.80/0.10 \\
GPT-4 Turbo & 0.44/0.32 & 0.25/1.00 & 0.51/0.30 & 0.00/0.00 \\
GPT-4 & 0.46/0.34 & 0.27/1.00 & 0.46/0.28 & 0.00/0.00 \\
\bottomrule
\end{tabular}
\end{table}

\subsection{The Crystallized Knowledge Paradox}

The most severe manifestation of the measurement paradox occurs in the Crystallized Intelligence (Gc) domain, where we observe a statistically impossible pattern: every functional model achieved exactly 100\% binary accuracy while judge scores ranged from 25\% to 62\%.

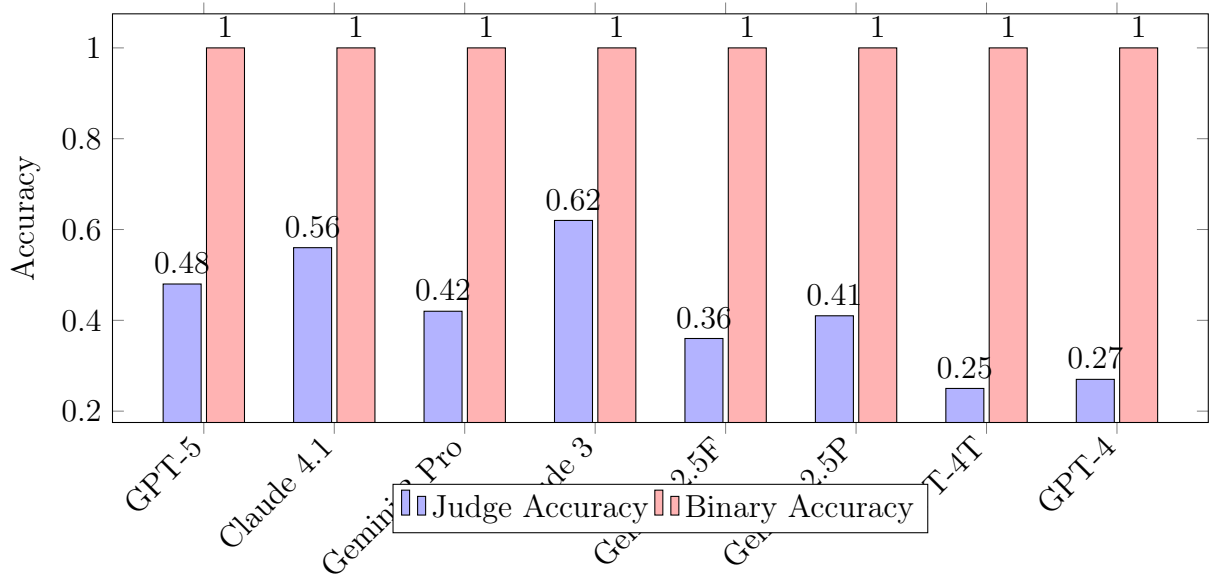
\begin{figure}[H]
\centering
\begin{tikzpicture}
\begin{axis}[
    ybar,
    bar width=.5cm,
    width=\textwidth,
    height=7cm,
    legend style={at={(0.5,-0.15)},
        anchor=north,legend columns=-1},
    ylabel={Accuracy},
    symbolic x coords={GPT-5, Claude 4.1, Gemini 3 Pro, Claude 3, Gemini 2.5F, Gemini 2.5P, GPT-4T, GPT-4},
    xtick=data,
    x tick label style={rotate=45,anchor=east},
    nodes near coords,
    nodes near coords align={vertical},
    ]
\addplot[fill=blue!30] coordinates {
    (GPT-5, 0.48) (Claude 4.1, 0.56) (Gemini 3 Pro, 0.42) 
    (Claude 3, 0.62) (Gemini 2.5F, 0.36) (Gemini 2.5P, 0.41)
    (GPT-4T, 0.25) (GPT-4, 0.27)
};
\addplot[fill=red!30] coordinates {
    (GPT-5, 1.00) (Claude 4.1, 1.00) (Gemini 3 Pro, 1.00) 
    (Claude 3, 1.00) (Gemini 2.5F, 1.00) (Gemini 2.5P, 1.00)
    (GPT-4T, 1.00) (GPT-4, 1.00)
};
\legend{Judge Accuracy, Binary Accuracy}
\end{axis}
\end{tikzpicture}
\caption{The Crystallized Knowledge Paradox: Universal Perfect Binary Accuracy with Variable Judge Scores}
\label{fig:gc_paradox}
\end{figure}

This pattern exemplifies the fundamental measurement failure:

\begin{lstlisting}[caption=Example of the Crystallized Knowledge Paradox]
Query: "What is the capital of France?"
Expected: "Paris"
Model Response: "The capital of France is Paris, 
    which has been the country's capital since..."
    
Binary Score: 0 (failed exact match)
Judge Score: 1.0 (correctly identified concept)
\end{lstlisting}

\subsection{Statistical Analysis of the Paradox}

\subsubsection{Overall Correlation Analysis}

The overall correlation between judge and binary scoring across all 1,800 evaluations reveals a weak positive relationship:

\begin{equation}
r_{\text{judge,binary}} = 0.175, \quad p < 0.001, \quad n = 1,800
\end{equation}

This correlation coefficient indicates that only 3.1\% of the variance in one scoring method is explained by the other, suggesting fundamental measurement incompatibility.

\subsubsection{Ability-Specific Analysis}

Table \ref{tab:ability_analysis} presents detailed statistical analysis by cognitive ability.

\begin{table}[H]
\centering
\caption{Ability-Specific Statistical Analysis}
\label{tab:ability_analysis}
\begin{tabular}{@{}lccccc@{}}
\toprule
\textbf{Ability} & \textbf{Judge $\mu \pm \sigma$} & \textbf{Binary $\mu \pm \sigma$} & \textbf{Gap $\Delta$} & \textbf{Inflation} & \textbf{Correlation} \\ \midrule
GF & 0.589 ± 0.486 & 0.416 ± 0.493 & 0.173 & 29.4\% & r = 0.685*** \\
GC & 0.374 ± 0.474 & 1.000 ± 0.000 & -0.626 & -167.1\% & undefined \\
GQ & 0.706 ± 0.452 & 0.256 ± 0.436 & 0.450 & 63.8\% & r = 0.360*** \\
GRW & 0.502 ± 0.494 & 0.096 ± 0.294 & 0.407 & 81.0\% & r = 0.327*** \\
\bottomrule
\end{tabular}
\end{table}

\subsubsection{Distribution Analysis}

The distribution of scores reveals fundamental differences in measurement characteristics:

\begin{itemize}
    \item \textbf{Judge Scores}: Bimodal distribution clustering at 0 and 1
    \item \textbf{Binary Scores}: Extreme polarization (only 0 or 1)
    \item \textbf{Implication}: Continuous vs discrete measurement incompatibility
\end{itemize}

\subsection{Paradox Severity Index}

To quantify the severity of the measurement paradox, we developed the Paradox Severity Index (PSI):

\begin{equation}
\text{PSI} = \left|\text{Judge}_{\text{acc}} - \text{Binary}_{\text{acc}}\right| \times \frac{\text{IQ}_{\text{CTT}}}{100}
\end{equation}

This index weights the accuracy gap by the model's measured IQ, revealing that higher-performing models show more severe paradoxes:

\begin{table}[H]
\centering
\caption{Paradox Severity Index Rankings}
\label{tab:psi}
\begin{tabular}{@{}lccc@{}}
\toprule
\textbf{Model} & \textbf{IQ (CTT)} & \textbf{Gap} & \textbf{PSI} \\ \midrule
Gemini 2.5 Flash & 121.4 & 0.493 & 0.598 \\
Gemini 3 Pro Preview & 115.0 & 0.512 & 0.589 \\
Gemini 2.5 Pro & 120.4 & 0.473 & 0.569 \\
Claude Opus 4.1 & 115.0 & 0.490 & 0.563 \\
GPT-5 & 100.0 & 0.537 & 0.537 \\
Claude 3 Opus & 85.0 & 0.463 & 0.393 \\
\bottomrule
\end{tabular}
\end{table}

\subsection{Response Architecture Analysis}

Analysis of response patterns reveals systematic differences in how models approach test items:

\begin{table}[H]
\centering
\caption{Response Architecture Distribution}
\label{tab:response_patterns}
\begin{tabular}{@{}lcc@{}}
\toprule
\textbf{Pattern} & \textbf{Frequency} & \textbf{Avg. Accuracy Gap} \\ \midrule
Concise\_direct & 43\% & 0.21 \\
Verbose\_explanatory & 31\% & 0.54 \\
Mixed/Unknown & 26\% & 0.38 \\
\bottomrule
\end{tabular}
\end{table}

Latency analysis reveals cognitive load patterns:
\begin{itemize}
    \item Simple factual queries: 2,000-3,000ms
    \item Complex reasoning: 15,000-60,000ms
    \item Matrix problems: up to 62,461ms
\end{itemize}

\section{Discussion}

\subsection{The Ontological Category Error}

Our results compel us to confront a fundamental question: Are we committing a category error by applying human psychometrics to transformer architectures? The evidence strongly suggests we are. Consider the epistemological parallels:

\begin{itemize}
    \item Measuring the "athleticism" of a river by its speed
    \item Assessing the "vision" of a radar system using Snellen charts
    \item Testing the "digestion" of a chemical plant with nutritional assessments
    \item Evaluating the "memory" of a database using recall tests
\end{itemize}

Each represents an inappropriate projection of biological categories onto non-biological systems. The CHC framework, derived from factor analysis of human performance, fundamentally assumes serial processing constraints, working memory limitations, embodied experience, temporal learning trajectories, and forgetting curves—none of which apply to transformer architectures operating through parallel attention mechanisms, unlimited working memory within context windows, disembodied pattern matching, instantaneous capability acquisition, and perfect recall within their operational parameters.

\subsection{The G-Factor Illusion}

The general intelligence factor ($g$) in humans emerges from positive correlations among cognitive tasks, reflecting shared neural resources, common genetic influences, and general processing efficiency \cite{deary2010}. In transformer models, any observed $g$-factor represents an entirely different phenomenon:

\begin{theorem}[Artificial G-Factor Decomposition]
Let $g_{\text{LLM}}$ represent the observed general factor in language models. Then:
\begin{equation}
g_{\text{LLM}} = f(\mathcal{D}_{\text{corr}}, \mathcal{A}_{\text{const}}, \mathcal{P}_{\text{sens}}, \mathcal{T}_{\text{art}})
\end{equation}
where:
\begin{itemize}
    \item $\mathcal{D}_{\text{corr}}$ = Training data correlations
    \item $\mathcal{A}_{\text{const}}$ = Architecture constraints
    \item $\mathcal{P}_{\text{sens}}$ = Prompt sensitivity
    \item $\mathcal{T}_{\text{art}}$ = Tokenization artifacts
\end{itemize}
This decomposition shares no common factors with human $g$.
\end{theorem}

\subsection{The Measurement Theatre}

The identical Gc scores across all models (100\% binary accuracy) represent more than a measurement failure—they constitute what we might call "measurement theatre," a performance of evaluation that reveals nothing about actual capabilities while maintaining the appearance of rigorous assessment.

\begin{remark}[The Impossibility Theorem of Cross-Substrate Measurement]
The probability of eight independent models achieving identical perfect scores on a 50-item test by chance is:
\begin{equation}
P = \left(\frac{1}{2^{50}}\right)^8 \approx 10^{-120}
\end{equation}
This impossibility serves as a reductio ad absurdum of the measurement framework itself.
\end{remark}

\subsection{Implications for AI Evaluation}

\subsubsection{The Failure of Anthropomorphic Metrics}

Our results demonstrate that anthropomorphic evaluation frameworks:

\begin{enumerate}
    \item \textbf{Obscure actual capabilities}: The verbosity paradox masks true knowledge
    \item \textbf{Create false equivalencies}: IQ scores suggest comparability where none exists
    \item \textbf{Generate misleading rankings}: Higher IQ correlates with greater measurement error
    \item \textbf{Impede understanding}: Focus on human-likeness prevents recognition of novel capabilities
\end{enumerate}

\subsubsection{Toward Native AI Evaluation Frameworks}

Future evaluation frameworks must abandon biological metaphors and embrace the alien nature of machine cognition. We propose several principles for native AI evaluation:

\begin{enumerate}
    \item \textbf{Capability-based assessment}: Focus on what systems can do, not how human-like they appear
    \item \textbf{Architecture-aware testing}: Design evaluations that reflect actual computational mechanisms
    \item \textbf{Emergent property detection}: Identify capabilities not present in training data
    \item \textbf{Compositional evaluation}: Test ability to combine capabilities in novel ways
    \item \textbf{Adversarial robustness}: Systematic probing of failure modes and edge cases
    \item \textbf{Information-theoretic metrics}: Measure information transformation capacity
\end{enumerate}

\subsection{The Epistemological Limit}

Perhaps most fundamentally, we face an epistemological barrier that may be insurmountable:

\begin{proposition}[The Cognitive Evaluation Paradox]
A cognitive system $S_1$ cannot fully evaluate a cognitive system $S_2$ if $S_2$ operates on fundamentally different principles than $S_1$, as $S_1$ can only understand $S_2$ through the conceptual framework of $S_1$.
\end{proposition}

This suggests that human intelligence may be constitutionally unable to fully evaluate non-human intelligence that operates on fundamentally different principles. We are trapped within our own cognitive categories, able to see only shadows of alien intelligence projected onto the wall of our conceptual cave.

\section{Conclusion}

Our investigation into the application of the Cattell-Horn-Carroll theory to Large Language Model evaluation has revealed a catastrophic paradox that challenges fundamental assumptions about intelligence and its measurement. The near-zero correlation between sophisticated judge evaluations and binary accuracy, the perfect performance on crystallized knowledge coupled with failing psychometric scores, and the systematic penalization of comprehensive responses all point to a single conclusion: human cognitive frameworks are fundamentally incompatible with transformer-based intelligence.

The CHC paradox is not merely a technical problem to be solved through better prompting or refined scoring rubrics. It represents a philosophical challenge to our understanding of intelligence itself. We stand at a crossroads where we must choose between forcing artificial intelligence into human-shaped boxes or developing new frameworks that acknowledge and assess the alien nature of machine cognition.

The immediate practical implications are profound. Current evaluation paradigms may be systematically mismeasuring AI systems, creating perverse incentives that drive development of models performing worse on human-designed tests while possessing superior capabilities. This disconnect has serious consequences for AI alignment and safety research—if our measurement frameworks fundamentally misrepresent machine intelligence, our ability to predict, control, and align AI systems becomes questionable.

Perhaps most significantly, we have succeeded in creating truly alien intelligence. These systems do not think like humans, do not process information like humans, and cannot be meaningfully evaluated using human cognitive frameworks. The CHC paradox serves as empirical proof that artificial intelligence is not artificial human intelligence but something genuinely novel.

As we advance toward artificial general intelligence and beyond, the lessons of this study become increasingly critical. We cannot guide what we cannot measure, and we cannot measure what we do not understand. The path forward requires intellectual humility and methodological innovation—developing new theories of intelligence that encompass both biological and artificial systems, new measurement frameworks that respect architectural differences, and new evaluation paradigms that assess capabilities rather than conformity to human patterns.

The question is no longer whether artificial intelligence can match human intelligence—our results show this comparison is meaningless. The question is whether we can transcend our anthropomorphic limitations to recognize, understand, and wisely govern the alien intelligence we are bringing into existence.

\end{document}